\documentclass[preprint,12pt]{elsarticle}

\usepackage{amssymb}
\usepackage{ragged2e}
\usepackage{array}
\usepackage{longtable}
\usepackage{hyperref}

\newcolumntype{P}[1]{>{\centering\arraybackslash}p{#1}}
\newcolumntype{M}[1]{>{\RaggedRight\arraybackslash}m{#1}}
\hypersetup{
    colorlinks=true,
    linkcolor=blue,
    filecolor=magenta,      
    urlcolor=cyan,
    }
    
\begin{document}

\begin{frontmatter}



\title{Integrating Bayesian methods with neural network--based model predictive control: a review}


\author[inst1]{Asli Karacelik\corref{cor1}}
\ead{asli.karacelik@ntnu.no}
\cortext[cor1]{Corresponding author.}
\affiliation[inst1]{organization={Department of Mechanical and Industrial Engineering, Norwegian University of Science and Technology}, 
            city={Trondheim},
            postcode={7491}, 
            country={Norway}}

\begin{abstract}
In this review, we assess the use of Bayesian methods in model predictive control (MPC), focusing on neural-network–based modeling, control design, and uncertainty quantification. We systematically analyze individual studies and how they are implemented in practice. While Bayesian approaches are increasingly adopted to capture and propagate uncertainty in MPC, reported gains in performance and robustness remain fragmented, with inconsistent baselines and limited reliability analyses. We therefore argue for standardized benchmarks, ablation studies, and transparent reporting to rigorously determine the effectiveness of Bayesian techniques for MPC. 
\end{abstract}

%\begin{keyword}
%keyword one \sep keyword two
%\end{keyword}

\end{frontmatter}


\section{Introduction}
\label{sec:sample1}
MPC employs the process model explicitly to obtain a control signal while minimizing an objective function \cite{2007_Camacho}. A model and current measurements predict future outputs and give an input-output relationship. Explicit use of the process model is advantageous in dealing with disturbances and constraints. Nonetheless, MPC has disadvantages. Its reliability depends on the process model. For highly nonlinear systems, it is not easy to obtain reasonably accurate models \cite{2007_Camacho}. Therefore, neural network systems are employed to acquire more reliable nonlinear models. We improve neural network models to have a better transient response, which adaptive controllers cannot achieve \cite{2016_Goodfellow}. Despite being more reliable, neural networks still have problems associated with uncertainty. Neural networks depend on statistical evaluation to quantify uncertainty \cite{2016_Goodfellow}. Relying on a statistical assessment is a weakness of neural network systems, as it is an estimate of uncertainty. We calibrate the neural network models for predictive uncertainty. Some uncertainty quantification methods are the single network deterministic, ensemble, and test-time augmentation methods \cite{Gawlikowski2021ASO}. Single network deterministic methods use deterministic parameters which perceived as true and accurate. In this method, a single network pass defines uncertainty quantification. Ensemble methods combine predictions of different single-network deterministic methods into one prediction. Test-time data augmentation is one of the simplest uncertainty estimation methods. Data augmentation means creating new data from existing data by modifying the existing data. We test all the samples to obtain a predictive distribution for uncertainty measurement. 

Prediction in Bayesian methods is stochastic, which employs probability distributions for model parameters. Therefore, there are different model weights for each prediction. We can categorize Bayesian methods according to their approach to the approximation of posterior probability. Prior probability updated with new information gives posterior probability. Posterior probability can be hard to calculate, so we need to infer the probability using different methods. These methods are variational inference, sampling methods, and Laplace approximation. Variational inference methods are deterministic methods that use a predetermined family of distributions (variational family) \cite{2019_Marcot}. Parametric distributions, such as the Gaussian distribution, define the predetermined family of distributions. Variational inference aims to adjust the parameters to obtain a parametric distribution of the approximated posterior distribution. Kullback-Leibler's (KL) divergence method measures the proximity of parametric and actual posterior distributions. However, we cannot minimize proximity using KL directly. KL is the difference between the log probability of observed data and the evidence lower bound. Log probability of observed data is the evidence, and KL should be greater than zero because it represents the distance (proximity) between parametric and posterior distributions. We need to minimize this distance to obtain a more accurate representation of the actual posterior distribution by optimizing ELBO, as the log observed data is a fixed value. For this reason, we call the term in the definition of KL an evidence lower bound. 

Stochastic variational inference (Monte-Carlo) is popular among variational inference methods. We call it stochastic because it does not use the entire training set to optimize ELBO. One of the successful methods in stochastic variational inference is Monte Carlo dropout. The motivation behind this method is to get a simple structure for a neural network and arrange it to get more accurate results outside the training region. Accuracy increases due to the prediction of uncertainty during test time. This method deactivates (dropout) some neurons randomly during training and test time and sets different pathways (layers) for the same input-output relationship. It is variational due to these different layers, and having different layers makes the system more robust because we have more than one way to obtain the result. 

Sampling methods (Monte Carlo Methods) do not have a parametric model for uncertainty approximation \cite{2021_Lyea}. Instead, it draws samples from a probability distribution to approximate the uncertainty. Markov Chain Monte Carlo (MMCC) sampling is the most preferred algorithm. MMCC is popular because of its efficiency in high-dimensional problems \cite{2021_Lyea}. The other algorithms are rejection sampling, importance sampling, and particle filtering. 

This review presents the applications of Bayesian methods using neural networks in model predictive control. First, we discuss the strategies of each paper and classify them based on the uncertainty model (variational, sampling, Laplace approximation) they use. Finally, we want to understand how well these uncertainty methods predict under uncertainty.

\section{Review Method}
\begin{figure}[htp]
    \centering
    \includegraphics[width=10cm]{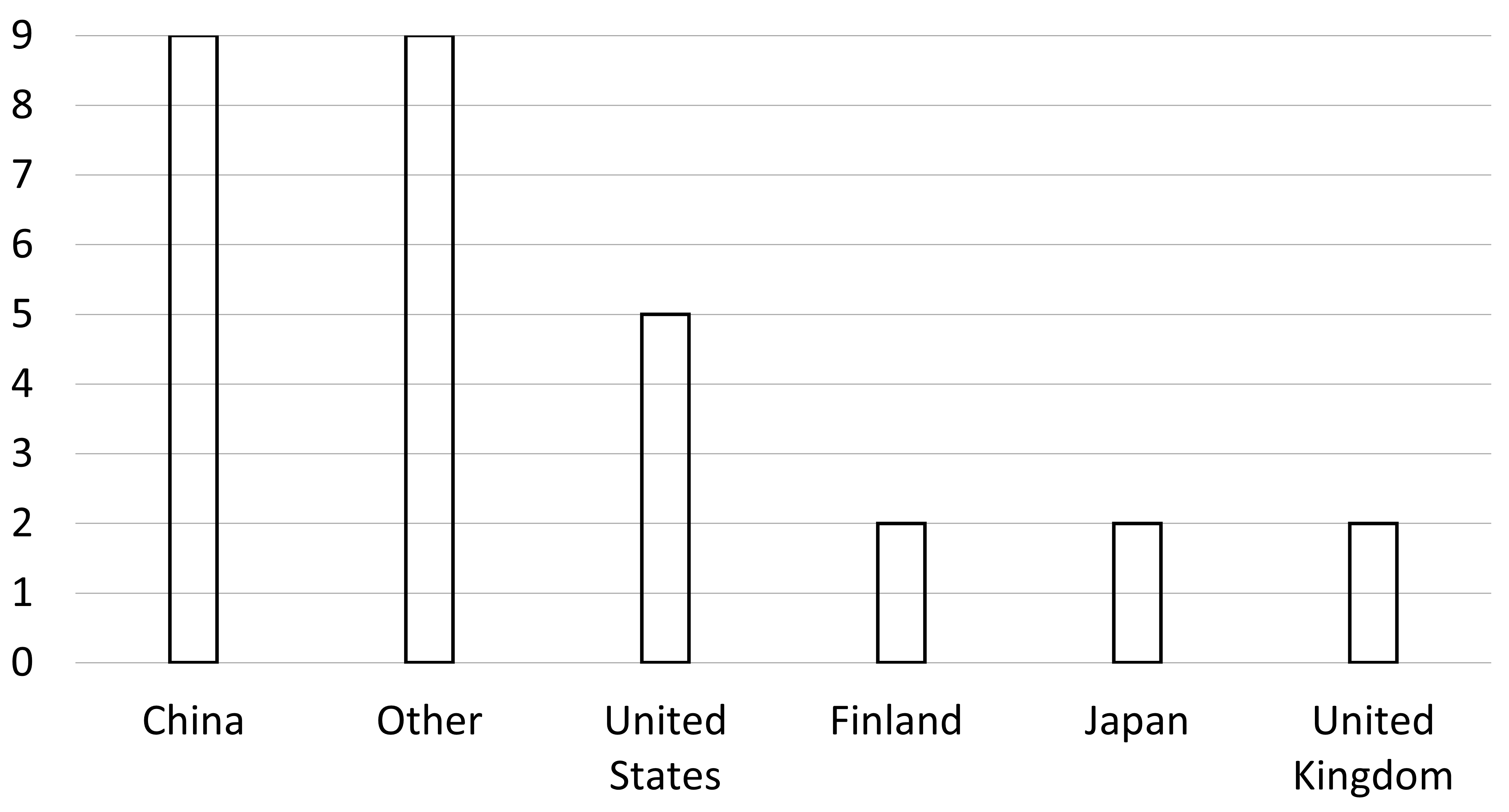}
    \caption{Number of articles by country}
    \label{fig:1}
\end{figure}

\begin{figure}[htp]
    \centering
    \includegraphics[width=10cm]{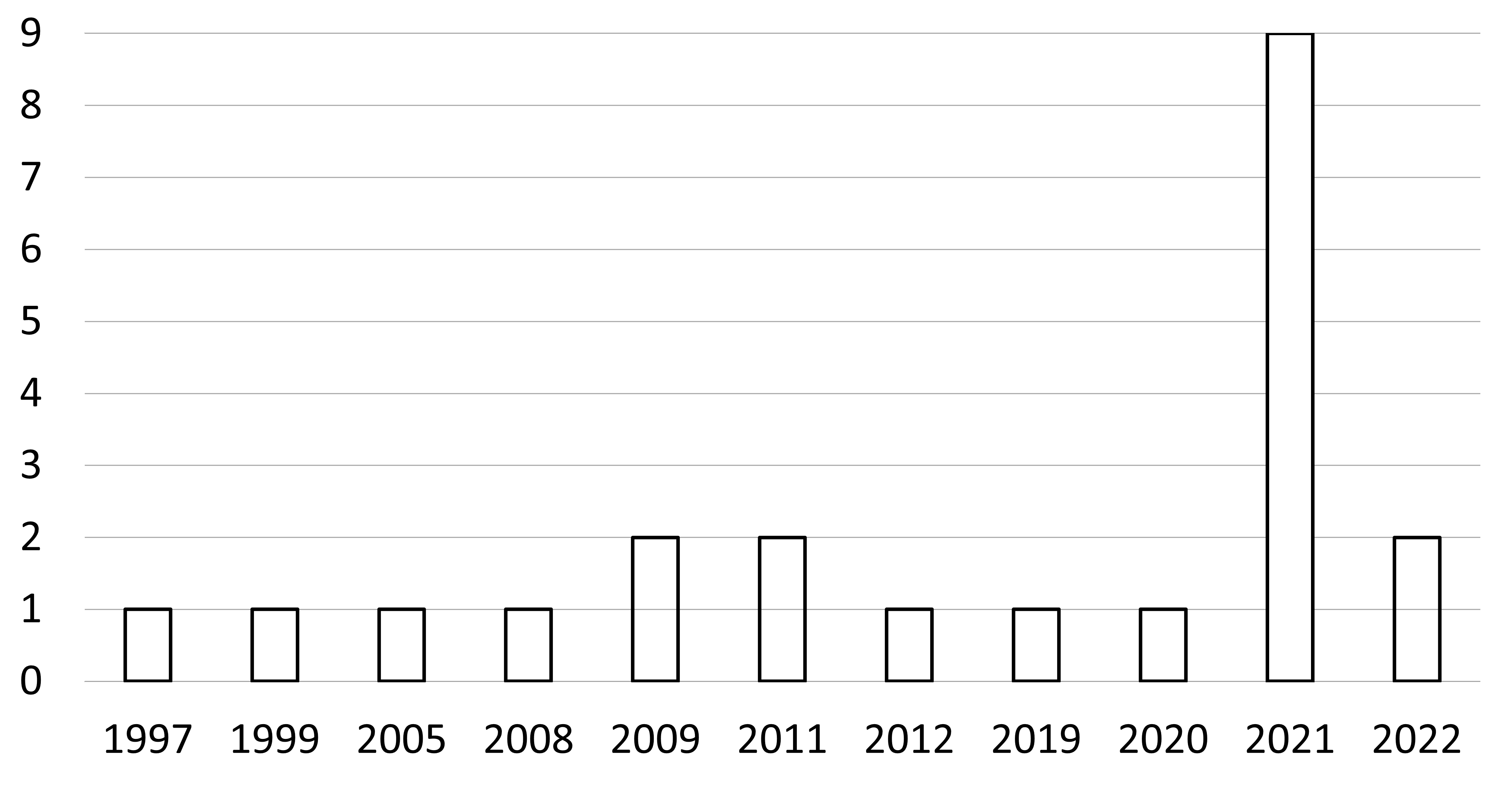}
    \caption{Number of articles by year}
    \label{fig:2}
\end{figure}

We introduced the keywords in the Scopus (Elsevier) database and presented the methods suggested by the articles. For this reason, we first scanned the abstracts and read the relevant articles in detail. We only reviewed articles and conference proceedings in the English language. \autoref{tab: search} shows the search results, and \autoref{fig:1} and \autoref{fig:2} demonstrate the number of articles by country and number of articles by year. 

\begin{table}[h!]
    \caption{Search Results}
    \label{tab: search}
 
        \begin{tabular}{M{9cm} M{4cm}}
  
        \textbf{Keywords}&\textbf{Number of Articles}\\ \hline
            TITLE-ABS-KEY ( bayesian  "model predictive control" )  AND  ( LIMIT-TO ( LANGUAGE ,  "English" ) )
             & 172\\  \hline
             TITLE-ABS-KEY ( bayesian  AND neural  AND network  AND  "model predictive control" ) &28\\ \hline
        TITLE-ABS-KEY ( bayesian  AND neural  AND network  AND  "model predictive control" )  AND  ( LIMIT-TO ( DOCTYPE ,  "cp" )  OR  LIMIT-TO ( DOCTYPE ,  "ar" ) )  AND  ( LIMIT-TO ( LANGUAGE ,  "English" ) ) &  22\\ \hline
      
           \end{tabular} 
\end{table}

\section{Review}

\subsection{Application Areas}
\begin{figure}[htp]
    \centering
    \includegraphics[width=10cm]{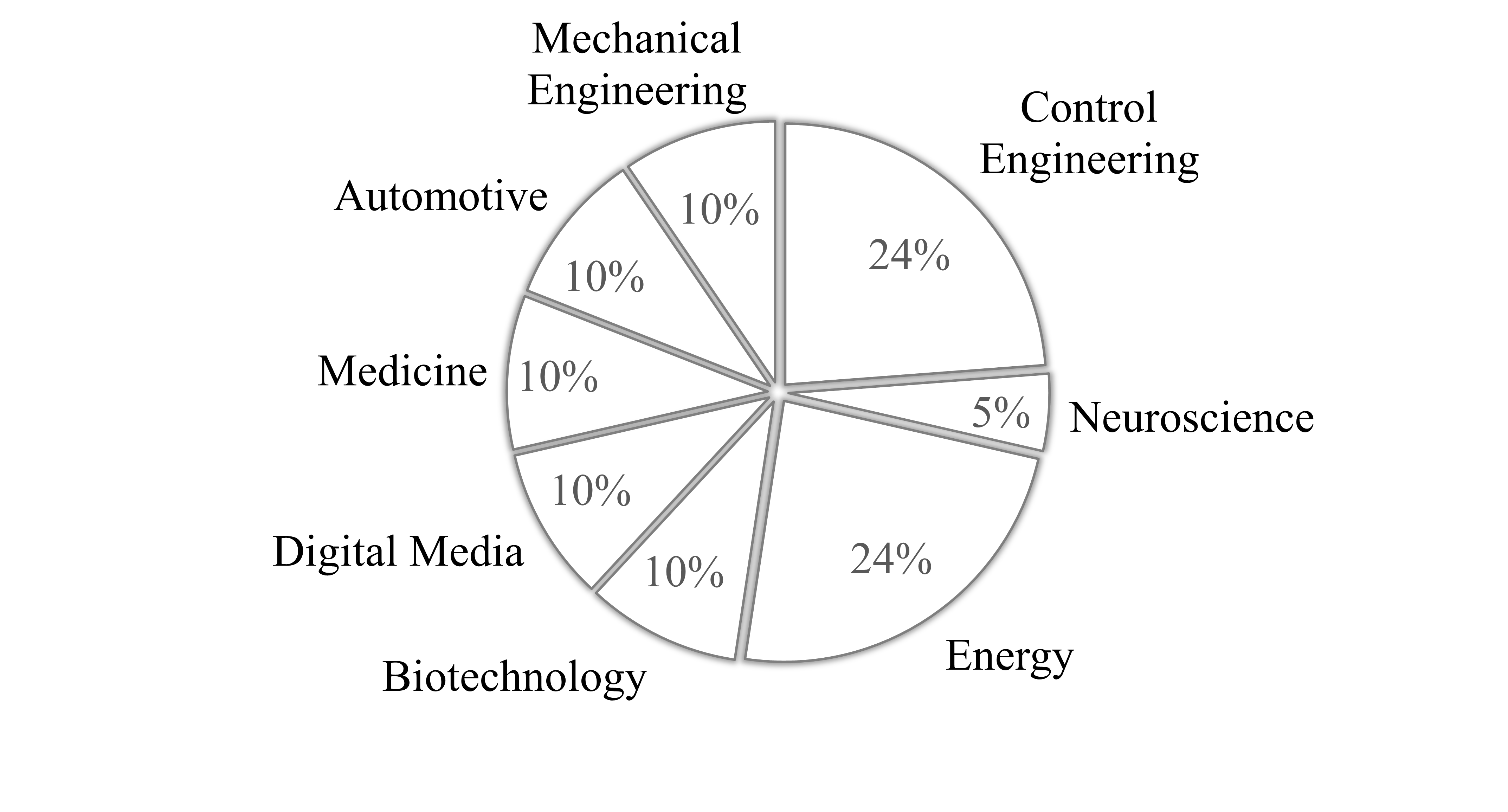}
    \caption{Percentage share of application areas}
    \label{fig:3}
\end{figure}

We use neural networks when it is difficult to describe the behavior of a system by a mathematical model. And the behavior of most systems is quite complex. Therefore, neural networks have a wide variety of application areas. \autoref{tab: application}, and \autoref{fig:3} show application areas and the percentage share for Bayesian methods integrated with neural network-based MPC, respectively.

\renewcommand{\arraystretch}{1.5}

\begin{longtable}{m{5.3cm}m{6.7cm}} 

  \caption{Application areas}
    \label{tab: application}\\
  
  \hline 
 \multicolumn{1}{m{5cm}}{\textbf{Author}}&
   \multicolumn{1}{m{7cm}}{\textbf{Application area}} 
  \\ \hline 
\endfirsthead

\multicolumn{2}{m{12cm}}
{{\bfseries \tablename\ \thetable{} -- continued from previous page}} \\ \hline
 \multicolumn{1}{m{5cm}}{\textbf{Author}}& \multicolumn{1}{m{7cm}}{\textbf{Application areas}}\\ \hline 
\endhead
\hline \multicolumn{2}{m{12cm}} {{Continued on next page}} \\ \hline
\endfoot
\hline
\endlastfoot
Ye and Ni (1997) \cite{1997_Ye} & Circulating fluidized bed boilers\\ \hline
Ye and Ni (1999) \cite{1999_Ye}& Circulating fluidized bed boilers\\ \hline
Raiko and Tornio (2005) \cite{2005_Raiko} & Cart-pole swing-up\\ \hline
Liu and Fang (2008) \cite{2008_Liu} & Hydraulic turbine\\ \hline
Cho et al. (2009) \cite{2009_Cho} & Inverted pendulum system\\ \hline
Raiko and Tornio (2009) \cite{2009_Raiko} & Cart-pole swing-up\\ \hline
Cho and Fadali (2011) \cite{2011_Cho} & Inverted pendulum system\\ \hline
Wei and Zhu (2011) \cite{2011_Wei} & Variable air volume (VAV) ventilation system\\ \hline
Wei and Liu (2012) \cite{2012_Wei} & Multi-zone VAV air conditioning system\\ \hline
Peng et al. (2019) \cite{2019_Peng} & Ultra-supercritical thermal power units of automatic generation control \\ \hline
Okada et al.(2020) \cite{2020_Okada} & DeepMind ControlSuite Simulation	Tasks: Ball-in-cup, catch finger, spin cheetah, run walker\\ \hline
Bonzanini et al. (2021) \cite{2021_Bonzanini} &	Cold atmospheric plasma jet \\ \hline
Chen et al. (2021) \cite{2021_Chen} & Artificial pancreas\\ \hline
Cursi et al. (2021) \cite{2021_Cursi} & Tendon-driven surgical robot\\ \hline
Fukami and Omori (2021) \cite{2021_Fukami} & A neural system of the brain \\ \hline

Jiang et al.(2021) \cite{2021_Jiang} & Preventing lateral collision of cars \\ \hline
Kan et al. (2021) \cite{2021_Kan} &	Video streaming \\ \hline
Morabito et al. (2021) \cite{2021_Morabito} &	Repetitive biotechnological process \\ \hline
Wang et al. (2021) \cite{2021_Wang} & Video streaming \\ \hline
Alquennah et al. (2022) \cite{2022_Alquennah} & Cross-over switches cell (CSC) inverter \\ \hline
Hu et al. (2022) \cite{2021_Hu} & Multi-range speed prediction \\ \hline
\end{longtable}

Below, we explained application areas for those who are not familiar with the subject areas:

• \textbf{Automatic generation control}: Power plants generate electricity from energy resources such as thermal and hydraulic energy. These energy resources turn into mechanical energy in generators. According to the online control performance standard, automatic generation control (AGC) regulates the power output from generators in different power plants. Then, the AGC grid system distributes the power from various power plants based on changes in load (required energy). There must be a balance between load and generation. Otherwise, a power cut will happen. Otherwise, there will be a power outage. We monitor this balance by frequency. When the load is higher, the frequency decreases and vice versa.

• \textbf{Circulating fluidized bed boilers}: Some thermal power plants utilize circulating fluidized bed boilers to produce steam by burning solid fuels. Air carries fine solid particles through the furnace, and the solid particles return to the furnace with the aid of a solid-gas cyclone. This process increases combustion efficiency. The produced steam is then used in turbines to produce mechanical energy. Furthermore, ultra-critical thermal power plants have high critical temperature and pressure operating points. Therefore, efficiency is higher than in conventional thermal power plants. 

• \textbf{Hydraulic turbines}: Hydroelectric power plants employ hydraulic turbines to generate power. They convert falling water energy into mechanical energy.  

• \textbf{Inverted pendulum}: An inverted pendulum is a standard system for control engineers to test their control strategies. However, the pendulum is unstable, and it falls without any support. The pendulum is usually attached to a cart, called a cart-pole swing-up. The cart tries to balance the pendulum upwards by sliding it left and right. The aim is to keep the pendulum suspended upward with a control strategy.

• \textbf{Variable air volume ventilation system}: Variable air volume (VAV) and constant air volume (CAV) systems are a type of air conditioning system. VAV changes airflow to keep the temperature steady, while CAV varies the temperature to supply constant airflow. As a result, VAV requires a lower fan speed, creates less noise, and provides accurate temperature control, increasing air conditioning systems' efficiency. In addition, multi-zone systems increase efficiency more by offering different climate conditions for each area. Thus, it is possible to set a different temperature for areas we do not use much.

• \textbf{Cold atmospheric plasma jet}: Plasma is one of the states of matter, and it happens when we give more energy to a gas. Energy rise in the gas creates charged particles and a thermodynamically unstable environment. We can provide this energy with heating or electrical power. These particles, for instance, ions, ultraviolet photons, and electrons, make the plasma electrically conductive and chemically reactive. Cold plasmas are near body temperature, leading to a new research area called plasma medicine. In addition to medicine, biomaterial processing also benefits from this technology. We can generate atmospheric cold plasma jets by creating an electric field with electrodes. 

• \textbf{Artificial pancreas}: Artificial pancreas controls insulin dosage so that the blood glucose remains around the safe range. In hypoglycemia, blood sugar is lower than usual, and blood sugar is higher in hyperglycemia. Hypoglycemia may lead to stroke, blindness, and kidney failure. Also, hyperglycemia can cause unconsciousness and death. Therefore, it is necessary to keep blood glucose in the normal range. Traditional artificial pancreas systems use simple linear models with MPC because medical devices cannot perform complex nonlinear and nonconvex online optimization.

• \textbf{Tendon-driven surgical robots}: Tendon-driven surgical robots mimic the movement system of a human. The human movement system comprises the muscle (biological motor)-tendon-bone relationship. In this relationship, muscle fibers not only transmit force to the tendons but also a harmony between them ensures movement. In tendon-driven surgical robots, these tendons are wires, and equipment like pulleys provides connections between them. As a result of these connections, we need fewer actuators; additionally, they offer safety and precision but also complicate the system. Because of this complexity, we must use advanced control mechanisms such as MPC. 

• \textbf{Repetitive process}: A repetitive process means rapid sequential production. This process is practical when we produce the same or similar products sequentially over a long time. We use the same procedure for the goods produced by this process. For example, batch or fed-batch processes can be repetitive. In batch processes, we give the raw materials at the beginning and wait for the production. In the fed-batch processes, we provide raw materials continuously or gradually during the process. We commonly use fed-batch processes in biotechnology called fed-batch cultivation. 

• \textbf{Crossover switches cell inverter}: The crossover switches cell inverter is a multilevel inverter topology that employs a minimum direct current (DC) source and switches while providing maximum voltage levels. Inverters convert DC into alternating current (AC). DC offers one direction for the flow of electric charge, while AC constantly changes direction and forms a sinusoidal form. Therefore, this conversion causes some distortion called total harmonic distortion. Voltage levels consist of different combinations of capacitor and DC source voltages. We introduce different voltage levels to approach a sinusoidal shape to reduce this harmonic distortion. For example, if we call the capacitor and DC source voltage V1 and V2, these combinations are as follows for nine levels: V1, V2, V1+V2, 0, -V2, V1-V2, -V1+V2, -V1, -V1-V2. The higher the number of levels, the smoother the sinusoidal waveform, but it also adds complexity and cost to the system. Consequently, a multilevel inverter is necessary to obtain a sinusoidal waveform. 

\subsection{Neural network models}

Neural networks and validation methods of the reviewed studies are shown in \autoref{tab: NN}.

\begin{longtable}{M{3cm}M{5cm}M{4cm}} 
  \caption{Neural network models}
    \label{tab: NN}\\ \hline 
  
  \multicolumn{1}{M{3cm}}{\textbf{Author}} & \multicolumn{1}{M{5cm}}{\textbf{Neural network model}} & \multicolumn{1}{M{4cm}}{\textbf{Validation method}} \\ \hline
\endfirsthead

\multicolumn{3}{M{12cm}}%
{{\bfseries \tablename\ \thetable{} -- continued from previous page}} \\ \hline
 \multicolumn{1}{M{3cm}}{\textbf{Author}}& 
 \multicolumn{1}{M{5cm}}{\textbf{Neural network model}}& \multicolumn{1}{M{4cm}}{\textbf{Validation method}} \\ \hline 
\endhead

\hline \multicolumn{3}{M{12cm}} {{Continued on next page}} \\ \hline
\endfoot
\endlastfoot

Ye and Ni (1997) \cite{1997_Ye} & • Bayesian-Gaussian neural network \break
 • Back Propagation neural network & LABVIEW dynamic model \\ \hline
Ye and Ni (1999) \cite{1999_Ye}  & • BGNN \break • BPNN & Semi-industrial laboratory experiment \\ \hline
Raiko and Tornio (2005) \cite{2005_Raiko} & Variational Bayesian learning &  Laboratory experiment \\ \hline
Liu and Fang (2008) \cite{2008_Liu} & BGNN & Industrial experiment\\ \hline
Peng et al. (2019) \cite{2019_Peng} & BNN rolling model &  • K-fold cross-validation \break • Comparative experiments of fuzzy algorithm modelling \cite{2014_Wu} \\ \hline
Wei and Zhu (2011) \cite{2011_Wei} & BNN & Laboratory experiment \\ \hline
Wei and Liu (2012) \cite{2012_Wei} & BNN & Laboratory experiment \\ \hline
Okada et al. (2020) \cite{2020_Okada} & Model-based reinforcement learning with Bayesian inference & DeepMind controlsuite simulation \\ \hline
Bonzanini et al. (2021) \cite{2021_Bonzanini} & • Deep neural network (DNN) trained with closed-loop data \break • Bayesian optimization for hyperparameters & Not mentioned \\ \hline
Chen et al. (2021) \cite{2021_Chen} & • Imitation learning method exploiting BNN \break • Deep recurrent neural \break network \break • Behavioral cloning & Virtual patients \\ \hline
Cursi et al. (2021) \cite{2021_Cursi} & BNN & Virtual robot experimentation platform (CoppeliaSim) \\ \hline
Fukami and Omori (2021) \cite{2021_Fukami}& • Sequential Monte Carlo method & Not mentioned \\ \hline
Jiang et al. (2021) \cite{2021_Jiang} & Bayesian regularized artificial NN & PreScan simulation \\ \hline
Hu et al. (2021) \cite{2021_Hu} & • Feed-forward neural  network \break • Bayesian neural network & Microscopic traffic simulation software (VISSIM)\\ \hline
Kan et al. (2021) \cite{2021_Kan} & • Convolutional NN point estimate \break • BNN point estimate & • Three real network trace data sets \break • Tenfold cross-validation method \\ \hline
Morabito et al. (2021) \cite{2021_Morabito} & • GPs \break • Bayesian inference of risk function & Laboratory experiment \\ \hline
Wang et al. (2021) \cite{2021_Wang} & • BNN \break • Backprop’s Bayes method & Two public datasets (FCC and 3G/HSDPA) \\ \hline

\end{longtable}
\sloppy
We used one source for each paragraph in this section unless stated otherwise.

Ye and Ni \cite{1997_Ye} have shown that the prediction results of BPNN are slightly better than BGNN for both static and transient performance, but BGNN requires less training time and is a simple method due to its self-tuning algorithm. On the other hand, BPNN requires a long training period due to the trial-and-error approach in parameter settings and topology. After this study, Ye and Ni \cite{1999_Ye} compared BPNN and BGNN, changing the algorithm in \cite{1997_Ye} such that BGNN captures the dynamics shift of the process for semi-industrial CFB. In this study, the authors performed online learning and found that BGNN gives better model prediction results. However, they only developed the neural network model for the MPC; they did not apply the MPC in these studies. Instead, they would use MPC in another study, but later, Reh and Ye \cite{2000_Reh} performed online prediction and optimization for semi-industrial CFB instead of applying MPC. 

Raiko and Tornio \cite{2005_Raiko} constructed a nonlinear hidden state-space model of a cart-pole system using variational Bayesian learning. In hidden state-space models, we don't know the relationship between state and observation; therefore, we use data to reveal this relationship. Nonlinear dynamic factor analysis \cite{2002_Valpola} provides system identification. This analysis uses multilayer perceptron networks for nonlinear mapping. 

Liu and Fang \cite{2008_Liu} proposed a BGNN method, which consists of an online application, offline learning, and a self-tuning process. The BGNN identifies the cost function used in the control algorithm. The authors emphasized that the BGNN prediction model matches the actual model. 

Cho et al. \cite{2009_Cho} developed a new online control method for a network control system combining reset control, nominal control, and neural networks independently. The online control structure involves a single-layer neural network for online learning. A perceptron algorithm, a linear machine learning algorithm, constructs this single-layer neural network. The authors aim to mitigate the time delay effect caused by network systems. Neural network and reset control lessen the system error due to time delay. The neural predictive control deploys a dynamic Bayesian network to predict error signals. Moreover, the Ubiquitous Sensor Network provides wireless communication for this networked control system. Later, Cho and Fadali \cite{2011_Cho} applied the same neural network approach to an inverted pendulum system with numerical simulations and experiments using a dc motor load.

Peng et al. \cite{2019_Peng} presented a Bayesian neural network model (BNN) of two coordinated systems in conventional ultra-supercritical thermal power units. The authors proposed a Bayesian neural network model with 92\% accuracy obtained from K-fold cross-validation. The input layer consists of steam turbine valve opening, water supply, and coal supply; the output layer involves actual power, intermediate point temperature, and mainstream pressure. When the authors compared the model with the fuzzy model offered by Wu et al. \cite{2014_Wu}, they found that the BNN provides better accuracy and faster convergence. 

Okada et al. \cite{2020_Okada} exploit Bayesian inference integrated with PlaNet (PlaNet-Bayes) for both incomplete dynamic models and optimal trajectories. The PlaNet is a deep planning network for reinforcement learning, and optimal trajectories define variational inference in MPC. The authors used a probabilistic action ensemble with trajectory sampling to include multimodel uncertainty. Multimodal learning combines different resources to gather information. In this study, multimodel uncertainty involves model and action uncertainties. The authors compared these uncertainties individually and together. The results indicate that multimodal Bayesian inference implemented in TensorFlow improves the asymptotic performance of the deep planning network.

Bonzanini et al. \cite{2021_Bonzanini} introduced a GP model for plant-model mismatch caused by uncertainties in state and inputs. The GP model defines plant-model mismatch in real-time and applies the scenario tree online. However, integrating GP predictions into optimal control problems (OCP) causes an increase in computational complexity. To reduce this complexity, a deep neural network (DNN) trained with closed-loop data estimates the LB-msMPC control law. The Bayesian optimization method determines optimal hyperparameters of the DNN by taking advantage of the bayesopt command in MATLAB. The DNN activation is the function ReLU, commonly used in regression tasks.

Cursi et al. \cite{2021_Cursi} proposed a BNN model for a tendon-driven surgical robot. The robot aims to follow the circle and square shapes in this study. The authors apply a Gaussian approximation with the Kullback-Leibler divergence method to obtain variational inference. Three neural network models form the kinematic model of the robot. In the initial kinematic model, the Denavit-Hartenberg convention defines the positions of arm links and joints of the robot. However, the simulation and kinematic models have different arm links. The authors chose the swish activation function, which unites the properties of ReLU and the sigmoid function to provide continuity in the derivatives. 
 
Fukami and Omori \cite{2021_Fukami} developed a data-driven method to estimate model parameters and the state of neural dynamics in the brain. The authors preferred to use the Morris-Lecar model to represent the biological neuron model. First, the stochastic expectation-maximization (EM) method estimates the parameters in this model. Then, the Sequential Monte Carlo method estimates the latent variables. Latent variables are not measured directly. Instead, observable variables predict latent variables using a mathematical model. The observed value is the membrane potential, and the Gaussian distribution defines the noise. Membrane potential is the electrical potential (voltage) gradient between the inside and outside of a biological cell. 

Jiang et al. \cite{2021_Jiang} implemented penalties on the shared control system, which provides a smooth transition between an intelligent driving system and human driving. MPC performs the dynamic optimization of this shared control policy. MATLAB quadratic and nonlinear programming solvers, quadprog and fmincon, offer a solution to this optimization problem. The results indicate that MPC outperforms LQR in evaluating risky events, preventing dangerous situations, and ensuring smooth transitions. 

Chen et al. \cite{2021_Chen} quantified predicted uncertainty using Monte Carlo Dropout, a Bayesian interpretation. State estimations are liable to error as they depend on the quality of measurements. Wrong state estimations cause severe problems, namely hypoglycemia and hyperglycemia. The proposed methods mitigate the input data shift between training and test data and eliminate state-estimated error. A deep recurrent neural network maps continuous glucose monitor measurements directly into the required insulin dosage. The authors indicate that the imitation learning method needs fewer supervision data and performs better than behavioral cloning. 
 
Kan et al. \cite{2021_Kan} developed BayesMPC, an adaptive bitrate algorithm, to improve the quality of experience for video streaming. The BayesMPC method estimates the statistical distribution of throughput, which is data transferred over a given period. The authors apply a variational approximation to find the probability distribution of network weights. The prior probability distribution is the Gaussian distribution, and the activation function is ReLU. Furthermore, this method alleviates the effect of the epistemic and aleatoric uncertainties by creating a confidence region for the future throughput. Epistemic uncertainty occurs due to insufficient training data, whereas aleatoric uncertainty occurs due to noise. With this study, the authors revealed the difference between point estimates and confidence regions. 

Morabito et al. \cite{2021_Morabito} combined an artificial neural network and first principles for a repetitive biotechnological process having model and measurement uncertainties. The process consists of two fed-batch reactors for two different products—one for product optimization and the other for profit optimization. The authors preferred to use the Adam algorithm for stochastic optimization. For the product optimization case, each layer holds a dropout rate of 20\%. The process model improves itself with every run by collecting measured data having Gaussian noise during cultivation. Furthermore, a risk function builds the balance between exploration and extrapolation. In this function, Bayesian inference with Gaussian process variances defines the uncertainties. GPyTorch and Pytorch train the Gaussian processes and neural networks, respectively. 
 
Wang et al. \cite{2021_Wang} developed an adaptive bitrate algorithm called 2prong to improve mobile video playback quality. Prediction of the amount of throughput is essential for a smooth video viewing experience. Unfortunately, stochastic network traffic and heavy-tailed network distribution make this prediction difficult. The authors combined neural networks and MPC to solve this problem for better video quality. First, the Bayesian neural network predicts throughput. Then, MPC optimizes the bitrate using throughput prediction and the previous bitrate. Finally, the authors compared the 2-prong algorithm with Pensieve (neural network), RobustMPC, rate-based, buffer-based, and BOLA (buffer-based, Lyapunov optimization) algorithms. The results show that the 2-prong algorithm provides 7.4\%-12.5\% better quality of experience than the Pensieve, which is the second-best algorithm. 

Alquennah et al.\cite{2022_Alquennah} applied a finite control set (FCS) MPC on a grid-connected crossover switches cell inverter that generates nine voltage levels. Inverter topology has a DC source and a capacitor voltage. The DC source (power supply) provides a single-phase voltage, meaning the electrical system contains one power line and one neutral line. The system creates different voltage levels by changing switches on and off and provides various states. There are limited switching states for power converters, so the control method is called FCS-MPC. The inputs to the Bayesian regularized feedforward learning method are weighting factors and reference values for currents; estimated THD and capacitor voltage error are the outputs. 

\subsection{Control methods}
Various control methods have been compared with MPC in some articles. \autoref{tab: control} shows the control methods compared and the improved MPC methods.

\begin{table}[h!]
 \caption{Control methods}
    \label{tab: control}
\begin{tabular}{M{3cm} M{6cm} M{4cm}}

\textbf{Author}&\textbf{Control method}&\textbf{Software}\\ \hline

Raiko and Tornio (2005) \cite{2005_Raiko} & • Direct Control \break • Optimistic inference control \break • NMPC & NFDA MATLAB package \\ \hline
Liu and Fang (2008) \cite{2008_Liu} & • MPC \break • PID & MATLAB Optimization Toolbox \\ \hline 
Okada et al. (2020) \cite{2020_Okada} & Variational inference MPC & Not mentioned \\ \hline
Bonzanini et al. (2021) \cite{2021_Bonzanini} & • Multi-stage MPC (msMPC) \break • Adaptive msMPC \break • Learning-based msMPC & MATLAB bayeopt CasADi (IPOPT solver) \\ \hline
Chen et al. (2021) \cite{2021_Chen} & • MPC with state estimation \break MPC with state information & MATLAB Interior-point algorithm \\ \hline
Cursi et al. (2021) \cite{2021_Cursi} & • Hierarchical MPC \break
• Pseudo inverse kinematic controller & MPC ACADO Toolkit \\ \hline
Hu et al. (2021) \cite{2021_Hu} & Multi-Horizon MPC & MATLAB Deep Learning Toolbox \\ \hline
Jiang et al. (2021) \cite{2021_Jiang} & • MPC \break • Linear quadratic regulator & MATLAB Deep Learning Toolbox MATLAB Interior-point algorithm \\ \hline
Kan et al. (2021) \cite{2021_Kan} & • BayesMPC \break
• RobustMPC & Not mentioned \\\hline
Morabito et al. (2021) \cite{2021_Morabito} & Shrinking horizon MPC & CasADi (IPOPT solver) \\ \hline

\end{tabular}

    \end{table}

Each paragraph below is based on one source only. 

Raiko and Tornio \cite{2005_Raiko} presented a direct control method where multilayer perceptron networks determine the future control input by superposing the inferred probability distribution and the expected value. The neural network operates as the controller in direct control methods. This method is successful in real-time but fails for pole stabilization. It also needs to learn policy mapping, which is hard to do well. OIC and NMPC are indirect control methods. OIC makes an optimistic guess, assuming some of the observations are true. It determines future control input using previous observations and control inputs. However, the two inference algorithms used in the experiments do not work well. It is also 100 times slower than the direct control. For NMPC, the purpose is to minimize a cost function within a predicted time. Current control input assumptions determine the probability distribution of the future states and observations. This method gives good results too, but it is 20 times slower than direct control. 

Liu and Fang \cite{2008_Liu} used the fmincon and fminunc commands of MATLAB Optimization Toolbox for the optimal control algorithm. The fmincon and fminunc are for constrained and unconstrained optimal control, respectively. The proposed method is simple and robust. This method shows that MPC with the BGNN predictive model outperforms the PID controller in fast and smooth response; the 2 Hz frequency step response results in a 6\% overshoot in the PID controller, while the presented method follows the step-change seamlessly.

Cho et al. \cite{2009_Cho} established a nominal control based on feedback linearization without time delay. The authors also a parallel reset control systems for offline and online control structures. Finally, they compared the results with another networked control system, a state feedback controller mentioned in \cite{2008_Tang}. The results show that the control system improves transient response and offers less overshoot and settling time (2.4 s). 

Cho and Fadali \cite{20011_Cho} implemented the same control method as Cho et al. with a DC motor load and found excellent impulse disturbance rejection and reference input tracking of a square wave. Furthermore, the controller system reduces settling time and overshoot caused by a square wave reference input.

Peng et al. \cite{2019_Peng} introduced intelligent predictive control based on control performance standard evaluation. The authors developed a novel control algorithm using a neural network for both the plant model and optimization. For optimization, the neural network rolling optimization model replaces the traditional rolling optimization model. Besides, the control algorithm also involves feedback correction. As a result, frequency inaccuracy and load overshoot are less than 0.2\% and 4\%, respectively. Finally, the authors compared the model with the fuzzy model offered by Wu et al. \cite{2014_Wu}. The proposed method responds faster, has high control robustness under pressure change, and the intermediate midpoint temperature is within an acceptable range. 

Bonzanini et al. \cite{2021_Bonzanini} offered an LB-msMPC method for complex, time-varying, and fast dynamics in the case of a plant-model mismatch. Worst-case scenario-based msMPC employs fixed uncertainty bounds for the scenario sets, whereas adaptive msMPC employs previous input and current state values to update the uncertainty bounds. The results show that LB-msMPC is superior in reaching the set point in the intended time. However, although adaptive msMPC outperforms msMPC, they do not get the setpoint during the planned time. 

Chen et al. \cite{2021_Chen} uncovered that MPC-driven adaptive stochastic policy keeps blood glucose in the normal range 8.4\%-11.75\% higher than MPC with the state estimate and 2.94\%–9.07\% higher than the deterministic policy.

Cursi et al. \citep{2021_Cursi} presented the Hi-MPC method for a robot that mimics the movement system of a human. Hi-MPC consists of two different MPC strategies. First, the primary MPC follows the chosen path. Next, another MPC minimizes the uncertainties in the model. Hi-MPC outperforms primary MPC in tracking shapes and meeting constraints. Also, the primary MPC fails to follow the frame path. Although the primary MPC has a lower mean square error for the circle path, it cannot meet the constraints. However, Hi-MPC has slightly higher errors in the z-axis direction and higher errors in initial positioning. Additionally, Hi-MPC has slower movement due to limitations and scaling factors.

Kan et al. \cite{2021_Kan} compared BayesMPC with RobustMPC, a convolutional neural network point estimate, and a BNN point estimate. Compared with RobustMPC and the convolutional neural network point estimate, the BNN point estimate increases the quality of experience by 20\% and reduces the mismatch probability by 7.8\% minimum. The mismatch that causes rebuffering is due to the difference between the video bitrate and the transfer rate (throughput). This study indicates that BayesMPC slightly performs better than the BNN point estimate regarding the quality of experience. Furthermore, the probability of a mismatch in BayesMPC decreases further by adjusting the confidence region compared to the BNN point estimation. Finally, BayesMPC is superior to other methods on generalization performance under untrained network conditions. 

Jiang et al.\cite{2021_Jiang} suggested a Bayesian regularized artificial neural network, which provides vehicle trajectories and a quantization function of risk assessment. The risk assessment reduces the computational intensity with an event-triggered control approach. Gauss-Newton approximation \cite{1997_DanForesee} and David MacKay's Bayesian method train the neural network. Three scenarios test this method for inexperienced driving, preventing rear-end collisions and lane-keeping in the PreScan simulation. 

Morabito et al. \cite{2021_Morabito} added a risk function to shrinking horizon MPC. The risk function builds a balance between exploration and extrapolation, and Bayesian inference from Gaussian processes defines the uncertainties in this risk function. The experiments show that the system converges at the fifth batch for low uncertainty and the seventh batch for high uncertainty. 

Alquennah et al.\cite{2022_Alquennah} built a cost function to minimize total harmonic distortion (THD) and regulate capacitor voltage to maintain nine voltage levels. According to the standard, total harmonic distortion should be less than 5\%. It also determines the switching state among 16 states to eliminate redundant switches. The method adjusts weighing factors in reaching the minimum cost value for different currents between 2A and 8A. The authors also applied fixed weighting factors and compared the results with the dynamic ones; the variable weighing factor technique decreases the THD of grid current by 4\%, 14\%, and 40\% for reference currents corresponding to 2A, 5.75A, and 8.25A. 

\section{Results}

Most of the articles we reviewed show that Bayesian methods give more robust and smoother results than other methods in the case of uncertainty. However, Ye and Ni stated that BPNN gives better results than BGNN, but BGNN is preferred because it is more practical. In addition, Cursi et al. \cite{2021_Cursi} claimed that Bayesian methods do not provide good results in some cases. In contrast, Bonzanini et al. \cite{2021_Bonzanini} and Gregori and Lightbody \cite{2012_Gregori} claim excellent results. We did not include the work of Gregori and Lightbody \cite{2012_Gregori} in this review because they used a predictive model in internal model control, but not in the MPC.

Gregori and Lightbody \cite{2012_Gregori} proposed a predictive model for internal model control (IMC) containing the Bayes Gaussian Process (GP) approach. The Bayesian approach inverts the GP model we cannot analytically reverse. The numerically inverted constraint is the estimated variance for the optimization problem.

Different results indicate that neural networks do not have a standard verification method. The structures of the systems are very different from each other, so it is difficult to apply a standardized validation method. For each system model, how much of data is sufficient to describe the behavior of that system varies. However, it is clear that the more data used, the better the result will be. As the operating speed of computers increases, we will mitigate some problems because working with a small amount of data increases the bias in neural network models.

We usually use variation inference in Bayesian methods because it is easy to implement. For this, most articles use this method to estimate uncertainty. However, accuracy depends on the parametric model and the similarity of the parametric model with the actual posterior probability. We build the parametric model based on the data we have so that the parametric model represents only the available data. Therefore, the accuracy of the data estimated under uncertainty is unreliable. However, sampling methods do not use parametric models, but accuracy still relies on data. If we do not use a parametric model, we also eliminate errors caused by data that does not fit the parametric model. However, sampling methods require large amounts of data to increase their accuracy, as they do not have a model to define uncertainty. Therefore, it requires high computational power. Only two authors, Chen et al. \cite{2021_Chen} and Qazani et al. \cite{2021_Qazani}, used sampling methods in their studies.

There is not enough information about the behavior of systems outside the training region, which means that Bayesian methods still cannot give reliable results in the case of uncertainty. Therefore, we should present more studies to understand our limitations in this regard and demonstrate the sensitivity of each system to the amount of training data in detail by applying different methods, like Monte Carlo dropout. 

\section{Conclusion}

Most of the studies we reviewed have suggested that Bayesian methods improve their systems. However, some studies have stated that the Bayesian methods do not give good results in some situations. These differences in inference are also based on the validation methods and the size of the training data. Most studies we examined did not test their models far outside the training region. Therefore, the reliability of Bayesian methods is uncertain. More studies should be done on this subject, and systems should be investigated based on data sensitivity to provide more reliable information in case of uncertainty.

\appendix
\section{Reviewed Articles}

\begin{longtable}{M{2cm}M{6cm}M{4cm}} 
  \caption{Reviewed Articles}
    \label{tab: articles}\\ \hline 
   \multicolumn{1}{M{2cm}}{\textbf{Author}}&
   \multicolumn{1}{M{6cm}}{\textbf{Title}}&
   \multicolumn{1}{M{4cm}}{\textbf{Journal}}\\ \hline 
\endfirsthead

\multicolumn{3}{M{12cm}}%
{{\bfseries \tablename\ \thetable{} -- continued from previous page}} \\ \hline
 \multicolumn{1}{M{2cm}}{\textbf{Author}}& \multicolumn{1}{M{6cm}}{\textbf{Title}}& \multicolumn{1}{M{4cm}}{\textbf{Journal}} \\ \hline 
\endhead

\hline \multicolumn{3}{M{12cm}} {{Continued on next page}} \\ \hline
\endfoot

\hline
\endlastfoot
Ye and Ni (1997) \cite{1997_Ye} & Static and transient performance prediction for CFB boilers using a Bayesian-Gaussian neural network & Journal of Thermal Science  \\ \hline
Ye and Ni  (1999) \cite{1999_Ye} & Neurocomputing  \\ \hline
Raiko and Tornio  (2005) \cite{2005_Raiko}  & Learning nonlinear state-space models for control & Proceedings of International Joint Conference on Neural Networks \\ \hline
Liu and Fang  (2008) \cite{2008_Liu}  & Predictive control strategy of hydraulic turbine turning system based on BGNN neural network & Lecture Notes in Computer Science \\ \hline
Peng et al.  (2019) \cite{2019_Peng} & Research on intelligent predictive AGC of a thermal power unit based on control performance standards & Energies \\ \hline
D. Wei and W.  (2011) \cite{2011_Wei} Zhu&Neural-network-based dynamic model of VAV systems&International Conference on Artificial Intelligence, Management Science and Electronic Commerce (AIMSEC)\\ \hline
D. Wei and X.  (2012) \cite{2012_Wei} Liu&Research on multi-zone VAV air conditioning system modeling
&Proceedings of the 10th World Congress on Intelligent Control and Automation \\ \hline
Okada et al.  (2020) \cite{2020_Okada} & PlaNet of the bayesians: Reconsidering and improving deep planning network by incorporating bayesian inference & IEEE/RSJ International Conference on Intelligent Robots and Systems (IROS) \\ \hline
Bonzanini et al. (2021) \cite{2021_Bonzanini} &Fast approximate learning-based multistage nonlinear model predictive control using Gaussian processes and deep neural networks&	Computers and Chemical Engineering\\ \hline
Cursi et al.  (2021) \cite{2021_Cursi} &Bayesian Neural Network Modeling and Hierarchical MPC for a Tendon-Driven Surgical Robot with Uncertainty Minimization&	IEEE Robotics and Automation Letters \\ \hline
Fukami and Omori  (2021) \cite{2021_Fukami} &Online Bayesian approach for estimation and control of neural system&IEEE 3rd Global Conference on Life Sciences and Technologies \\ \hline
Chen et al. (2021) \cite{2021_Chen} &MPC-guided Imitation Learning of Bayesian Neural Network Policies for the Artificial Pancreas&Conference on Decision and Control\\ \hline
Hu et al. (2021) \cite{2021_Hu} &A Multirange Vehicle Speed Prediction With Application to Model Predictive Control-Based Integrated Power and Thermal
Management of Connected Hybrid Electric Vehicles&Journal of Dynamic Systems\\ \hline
Jiang et al.  (2021) \cite{2021_Jiang} &Event-triggered shared lateral control for safe-maneuver of intelligent vehicles&Science China Information Sciences \\ \hline
Kan et al.  (2021) \cite{2021_Kan} &Uncertainty-Aware robust adaptive video streaming with bayesian neural network and model predictive control&Proceedings of the 2021 Workshop on Network and Operating System Support for Digital Audio and Video \\ \hline
Morabito et al. (2021) \cite{2021_Morabito} &Towards risk-aware machine learning supported model predictive control and open-loop optimization for repetitive processes&IFAC-PapersOnLine \\ \hline
Wang et al. (2021) \cite{2021_Wang} &2prong: Adaptive Video Streaming with DNN and MPC&Proceedings - 2021 17th International Conference on Mobility, Sensing and Networking, MSN 2021\\ \hline

\end{longtable}

 \bibliographystyle{elsarticle-num} 
 \bibliography{mybibliography}





% \end{thebibliography}
\end{document}